\documentclass[a4paper]{article}

\usepackage{INTERSPEECH2020}
\usepackage{multirow}
\usepackage{enumerate}
\usepackage{amsmath}
\usepackage{systeme}
\usepackage{graphicx}
\usepackage{paralist}
\usepackage{cite}

\usepackage{pgfplots}
\pgfplotsset{compat=1.13}
\usetikzlibrary{arrows.meta}
\usepackage{xcolor}
\definecolor{c1}{HTML}{b88b4d}
\definecolor{c2}{HTML}{79c561}
\definecolor{c3}{HTML}{ded94c}
\definecolor{c4}{HTML}{db79c0}
\definecolor{c5}{HTML}{ff362e}
\definecolor{c6}{HTML}{7982db}

\title{Group Gated Fusion on Attention-based Bidirectional Alignment for Multimodal Emotion Recognition}
\name{Pengfei Liu$^1$, Kun Li$^1$ and Helen Meng$^2$}
\address{
  $^1$SpeechX Limited, Shenzhen, China\\
  $^2$The Chinese University of Hong Kong, Hong Kong SAR, China}
\email{pfliu@speechx.cn, kli@speechx.cn, hmmeng@se.cuhk.edu.hk}

\begin{document}

\maketitle
\begin{abstract}
Emotion recognition is a challenging and actively-studied research area that plays a critical role in emotion-aware human-computer interaction systems. In a multimodal setting, temporal alignment between different modalities has not been well investigated yet. This paper presents a new model named as Gated Bidirectional Alignment Network (GBAN), which consists of an attention-based bidirectional alignment network over LSTM hidden states to explicitly capture the alignment relationship between speech and text, and a novel group gated fusion (GGF) layer to integrate the representations of different modalities.
We empirically show that the attention-aligned representations outperform the last-hidden-states of LSTM significantly, and the proposed GBAN model outperforms existing state-of-the-art multimodal approaches on the IEMOCAP dataset.
\end{abstract}


\section{Introduction}
Emotion recognition is a core component in any emotion-aware human-computer interaction system, such as an intelligent virtual assistant, and an affective spoken dialog system.
An emotion recognizer typically analyzes speech, text or images. For example, a speech emotion recognizer aims to identify the emotion carried in speech, often in terms of a set of emotion categories such as happy, angry, sad and neutral \cite{schuller2011recognising,han2014speech,satt2017efficient,ma2018emotion,wu2019speech}.
However, this is a nontrivial task because emotions are manifested in various factors such as conversational discourse, linguistic content and prosodic features \cite{tahon2015towards,barbulescu2017prosodic,aguilar2019multimodal}.
It is difficult to predict the emotion of an utterance based only on the acoustic features from speech, or only on the discrete word sequence in spoken text.
Furthermore, there is a lack of large-scale emotive datasets of speech or text, as these are costly and difficult to collect and label. One should therefore explore the use of multiple modalities for emotion recognition.

Multimodal emotion recognition has received a lot of attention in recent years \cite{aguilar2019multimodal,banziger2009emotion,cho2018deep,heusser2019bimodal,zadeh2017tensor,aldeneh2017pooling,yoon2018multimodal,poria2018multimodal,majumder2018multimodal,xu2019learning}. Modalities such as speech, text and/or images have been exploited for better recognition performance.
Existing approaches to multimodal learning are either \textit{early-fusion} which fuses low-level feature interactions between different modalities before making a prediction decision; or \textit{late-fusion} which models each modality independently and combines the decisions from each model \cite{aldeneh2017pooling}.
Recent early-fusion approaches have focused on different mathematical formulations to fuse acoustic and lexical features, such as multimodal pooling fusion \cite{aldeneh2017pooling}, modality hierarchical fusion \cite{majumder2018multimodal}, conversational neural network \cite{hazarika2018conversational}, word-level concatenation of acoustic and lexical features \cite{aguilar2019multimodal}, etc.
To capture inter-modality dynamics in multimodal sentiment analysis, Zadeh et al. \cite{zadeh2017tensor} introduces the Tensor Fusion Network (TFN) model using tensor fusion to explicitly aggregate unimodal, bimodal and trimodal interactions between different features.
Arevalo et al. \cite{arevalo2017gated} presents a Gated Multimodal Unit (GMU) model to find an intermediate representation based on a combination of features from different modalities, where it learns to decide how modalities influence the activation of the unit using multiplicative gates.
These methods are generally applicable in information fusion from different sources. However, they do not take into account the particular temporal correlation property of speech and text in an utterance, where a sequence of speech frames are aligned with a sequence of words temporally.

Given a sequence of speech frames, it is beneficial to know the corresponding words to learn a more discriminative representation for emotion recognition, and vice versa.
For example, consider a speaker expressing \textit{happiness} in the utterance ``\textit{That's great!}'', and \textit{anger} in ``\textit{That's unfair!}''.
An emotion recognizer should learn to pay more attention to the words \textit{great} and \textit{unfair} and their \textit{corresponding} speech frames, as there is complementary information across modalities for emotion recognition. This calls for a mechanism to capture the alignment between words and their speech frames.
Yoon et al. \cite{yoon2018multimodal} presents a deep dual recurrent neural network to encode information from speech and text sequences, and captures the alignment by a dot product attention to focus on specific words of a text sequence conditioned on speech for emotion classification. 
Similarly, Xu et al. \cite{xu2019learning} learns the alignment between speech and text using an additive attention, and adopts an LSTM layer fed with the concatenation of the aligned speech representation and the hidden-state text representation for emotion classification.
However, both approaches \cite{yoon2018multimodal,xu2019learning} consider only unidirectional alignment, and we will show empirically that bidirectional alignment brings better recognition performance. 
Furthermore, the concatenation-based fusion methods in \cite{yoon2018multimodal,xu2019learning} may not have sufficient expressive power to exploit the complementary information across modalities, which in turn reflects the need for more effective ways to fuse multiple representations.

In this paper, we propose a model named Gated Bidirectional Alignment Network (GBAN), which consists of a bidirectional attention-based alignment network to capture the alignment information between speech and text, and a novel group gated fusion (GGF) layer to automatically learn the contribution of each modality in determining the final emotion category.
We summarize the major contributions of this paper as follows:
\begin{enumerate}[(1)]
    \item Experimental results show that the proposed bidirectional alignment network leads to more discriminative representations of speech and text for emotion recognition;
    \item The proposed GGF method allows integration of multiple representations effectively, and obtains interpretable contribution weights of each modality in emotion recognition;
    \item The proposed GBAN model outperforms existing state-of-the-art multimodal approaches \cite{yoon2018multimodal,xu2019learning} using both speech and text on the IEMOCAP dataset \cite{busso2008iemocap}.
\end{enumerate}

\section{Gated Bidirectional Alignment Network}
\begin{figure}[htb]
  \centering
  \includegraphics[width=0.76\linewidth]{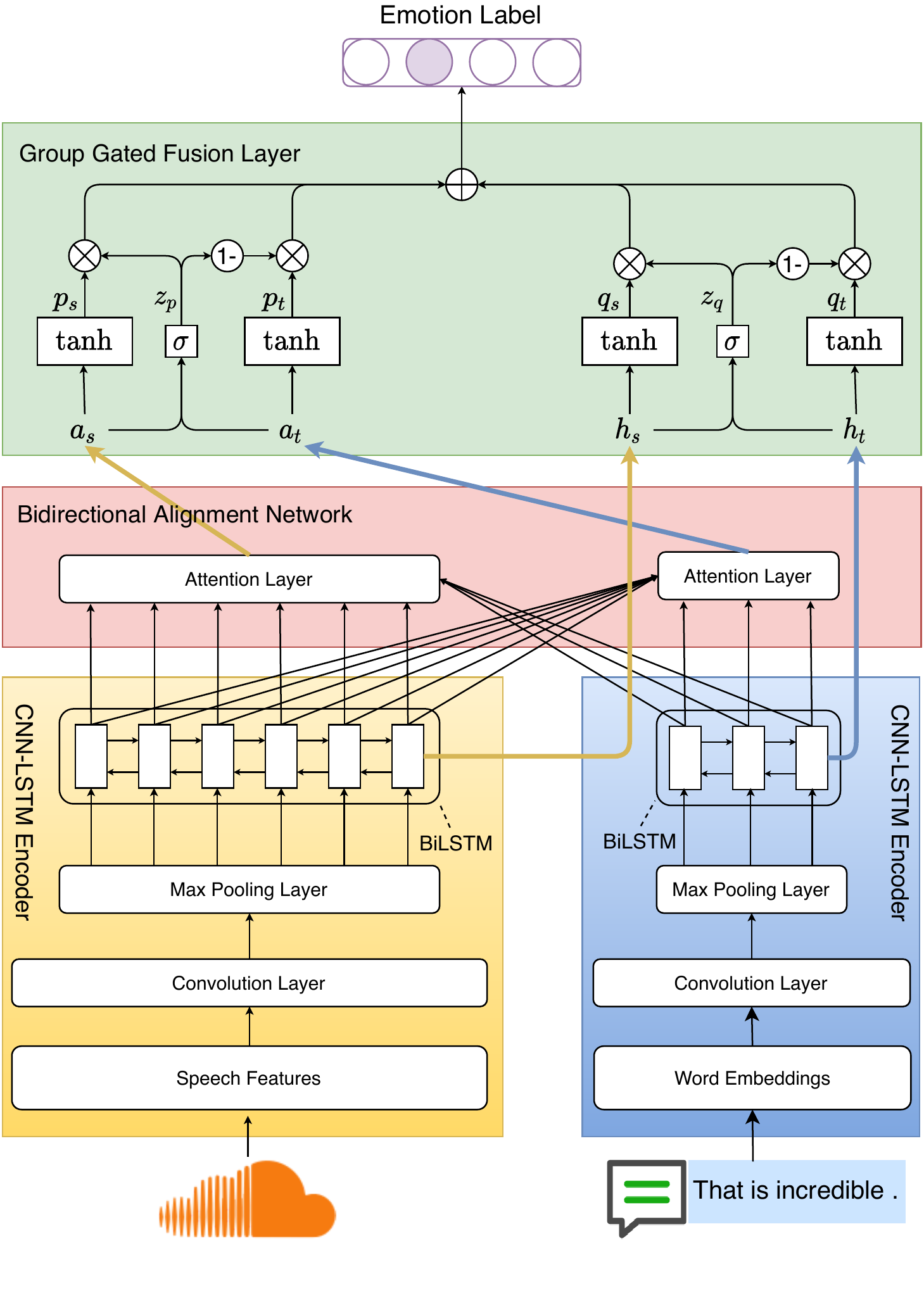}
  \vspace{-2em}
  \caption{The block diagram of the proposed GBAN model.}
  \label{fig:model}
\end{figure}

We propose an approach named Gated Bidirectional Alignment Network (GBAN) for multimodal emotion recognition.
As illustrated in Figure~\ref{fig:model}, the GBAN model consists of three major parts: 
\begin{inparaenum}[(1)]
    \item two separate CNN-LSTM encoders to extract features from speech and text respectively;
    \item an attention-based bidirectional alignment network to capture temporal correlations between speech and text; and
    \item a group gated fusion layer to learn the contribution of each representation automatically.
\end{inparaenum}

 \vspace{-1em}
\subsection{CNN-LSTM Encoders}
We adopt two separate CNN-LSTM encoders for speech and text respectively, where CNN layers extract local features and a bidirectional LSTM layer captures the global dependencies over time in either the speech acoustics or the text of an utterance.

\textbf{Speech Representation.}
The speech signal in an utterance is represented as a sequence of vectors $[x_1, \dots, x_N]$, where $N$ is the number of frames in the utterance.
First, the sequence is fed into a CNN layer, which consists of \texttt{convolution} and \texttt{max-pooling} operations, to obtain a sequence of pooled vectors $[p_1, \dots, p_K]$.
Then, a bidirectional LSTM layer follows the CNN layer to obtain a vector $s_i$ for each $i\in\{1,\dots,K\}$, which is concatenated from the forward and backward LSTMs.
Thus, the obtained sequence of $[s_1, \dots, s_K]$ keeps the temporal ordering for each $i\in\{1,\dots,K\}$, and will be used as the speech representation for alignment with the text representation.
In practice, multiple CNN layers and bidirectional LSTM layers are adopted to encode the low-level speech signals.
\begin{align}
    p_i &= CNN([x_1, \dots, x_N]), i\in\{1,\dots,K\} \\
    \overrightarrow{s_i} &= \overrightarrow{LSTM}(p_i), i\in\{1,\dots,K\} \\
    \overleftarrow{s_i} &= \overleftarrow{LSTM}(p_i), i\in\{1,\dots,K\} \\
    s_i &= [\overrightarrow{s_i}, \overleftarrow{s_{K-i+1}}], i\in\{1,\dots,K\}
\end{align}

\textbf{Text Representation.}
For text encoding, each sentence is represented as a sequence of vectors $[e_1, \dots, e_M]$, where $M$ is the number of words in a sentence, and $e_j$ is the word embeddings of the $j$th word.
Similar to the process in speech representation, a CNN layer is first used to obtain a sequence of pooled vectors $[q_1, \dots, q_L]$.
For each $j\in\{1,\dots,L\}$, a bidirectional LSTM layer follows to obtain the concatenation vector $t_j$ from the forward and backward LSTMs.
The sequence of $[t_1, \dots, t_L]$, which are essentially text features transformed from the sequence of $M$ words, keeps the temporal ordering for each $j\in\{1,\dots,L\}$, and will be used as the text representation for alignment with the speech representation of $[s_1, \dots, s_K]$.
\begin{align}
    q_j &= CNN([e_1, \dots, e_M]), j\in\{1,\dots,L\} \\
    \overrightarrow{t_j} &= \overrightarrow{LSTM}(q_j), j\in\{1,\dots,L\} \\
    \overleftarrow{t_j} &= \overleftarrow{LSTM}(q_j), j\in\{1,\dots,L\} \\
    t_j &= [\overrightarrow{t_j}, \overleftarrow{t_{L-j+1}}], j\in\{1,\dots,L\}
\end{align}

\subsection{Bidirectional Alignment Network}
An utterance is made up of a speech signal and a text sequence (transcribed by a person or a speech recognizer), where the speech and the text are sequentially correlated over time.
To capture their temporal correlations, we propose a bidirectional alignment network to learn the representation of one modality with the help of the other based on attention models.

\textbf{Attention-Aligned Speech Representation.}
Given the speech representation $[s_1, \dots, s_K]$ and the text representation $[t_1, \dots, t_L]$ obtained from the previous CNN-LSTM encoders for an utterance, we calculate the attention weight between the $i$th speech vector and the $j$th text vector as follows:
\begin{align}
    a_{j,i} &= \tanh{(s_i^\top t_j)} \\
    \alpha_{j,i} &= \frac{\exp^{a_{j,i}}}{\sum_{k=1}^K \exp^{a_{j,k}}} \\
    \widetilde{s_j} &= \sum_{k=1}^K\alpha_{j,k}s_k
\end{align}
where $\widetilde{s_j}$ is the weighted summation of the speech vectors, which is considered as an aligned speech vector corresponding to the $j$th text vector.
Finally, we apply an \texttt{average-pooling} operation on the sequence of aligned vectors to obtain the text-aligned speech representation $a_s = \texttt{average-pooling}([\widetilde{s_1}, \dots, \widetilde{s_L}])$.

\textbf{Attention-Aligned Text Representation.}
We apply a similar process to learn the attention-aligned text representation.
Given the text representation $[t_1, \dots, t_L]$ and the speech representation $[s_1, \dots, s_K]$ from the CNN-LSTM encoders for an utterance, we calculate the attention weight between the $j$th text vector and the $i$th speech vector as follows:

\begin{align}
    b_{i,j} &= \tanh(t_j^\top s_i) \\
    \beta_{i,j} &= \frac{\exp^{b_{i,j}}}{\sum_{l=1}^L \exp^{b_{i,l}}} \\
    \widetilde{t_i} &= \sum_{l=1}^L\beta_{i,l}t_l
\end{align}
where $\widetilde{t_i}$ is the weighted summation of the text vectors, which is considered as an aligned text vector corresponding to the $i$th speech vector.
Finally, we place an \texttt{average-pooling} layer to obtain the speech-aligned text representation $a_t=\texttt{average-pooling}([\widetilde{t_1}, \dots, \widetilde{t_K}])$.

\subsection{Group Gated Fusion Layer}
For the speech modality, we have obtained two representations: $a_s$ as the text-aligned speech representation and $h_s$ as the last hidden state of the BiLSTM layer. For the text modality, we have another two similar representations: $a_t$ and $h_t$.
To exploit these representations, we design a novel group gated fusion (GGF) layer to learn the contribution of each representation automatically.
Considering different grouped learning processes, we make $a_s$ and $a_t$ as the first group, $h_s$ and $h_t$ as the second group and design the two separate gates corresponding to each group. 
The first gate controls contributions of the aligned representations $a_s$ and $a_t$, while the second gate controls contributions of the last hidden states $h_s$ and $h_t$.
The equations governing the GGF layer are as follows:
\begin{align}
    p_s &= \tanh({W_s^a a_s}) \\
    p_t &= \tanh({W_t^a a_t}) \\
    z_p &= \sigma ({W_z^p [a_s, a_t]}) \\
    q_s &= \tanh({W_s^h h_s}) \\
    q_t &= \tanh({W_t^h h_t}) \\
    z_q &= \sigma ({W_z^q [h_s, h_t]}) \\
    h &=\underbrace{z_p \odot p_s + (1-z_p) \odot p_t}_{\text{Group 1}} + \underbrace{z_q \odot q_s + (1-z_q) \odot q_t}_{\text{Group 2}}
\end{align}
Here, $W_s^a, W_t^a, W_s^h, W_t^h$ are the weights for the  non-linear (i.e., \texttt{tanh}) transformations from $a_s, a_t, h_s, h_t$ respectively. $W_z^p, W_z^q$ are used to learn the contribution of each modality within each group.
$\sigma$ is the \texttt{sigmoid} function and $\odot$ means element-wise product.
$h$ is obtained by summing the gated representations of the two groups, and as the final representation for the following emotion classification layer.

\subsection{Emotion Classification Layer}\label{sec:softmax}
Given the obtained representation $h$, we first apply a fully-connected layer with rectified linear units (ReLUs) for non-linear transformation $g$, and use a \texttt{softmax} output layer to get $\tilde{y}$ for emotion classification of an utterance. 
The training objective $\mathcal{L}$ is to minimize the negative log-likelihood, where $N$ is the total samples in training, $C$ is the total number of emotion classes and $y_{i,c}=1$ if the ground-truth label is $c$ else 0.

\begin{align}
    g &= \texttt{ReLu}(W_g h) \\
    \tilde{y} &= \texttt{softmax}(W_e g + b) \\
    \mathcal{L} &= -\log \prod_{i=1}^N\sum_{c=1}^C y_{i,c} \log(\tilde{y}_{i,c})
\end{align}

\section{Experiments}

For model evaluation, we conducted 5-fold cross validation on the Interactive Emotional Dyadic Motion Capture (IEMOCAP) dataset \cite{busso2008iemocap}, which consists of five sessions with one male and one female speaker each.
We used 4 sessions as training set\footnote{We randomly select 5\% of the utterances as the validation set.} and the remaining session as testing set to ensure speaker independence.
To stay consistent with most previous investigations on IEMOCAP, we use the subset covering the four emotional categories of \textit{happy}, \textit{angry}, \textit{sad} and \textit{neutral}.

\subsection{Speech and Text Features}

We extract both mel-spectrogram and MFCCs as the acoustic features from the speech modality.
Each frame in the speech utterance corresponds to a feature vector consisting of 26-dimensional mel-spectrograms, 13-dimensional MFCCs and their first-order deltas, leading to a 52-dimensional vector. Following \cite{neumann2017attentive}, we set the maximal length of a speech utterance to 7.5s, with longer utterances cut at 7.5s and shorter ones padded with zeros. 

For each utterance in the IEMOCAP dataset, there is a corresponding human transcription, which can also be obtained by an automatic speech recognizer \cite{yoon2018multimodal,xu2019learning} in the deployed emotion recognition systems.
Both word-level and character-level embeddings can be used to represent the textual transcriptions.
We adopt word embeddings to represent each word within an utterance in the IEMOCAP dataset, and initialize the embeddings with the pre-trained 300-dimensional Glove vectors \cite{pennington2014glove}.

\subsection{Settings and Metrics}
We initialize all the network weights in the GBAN model with Xavier normal initializer \cite{glorot2010understanding}, and use the Adam \cite{adam2015kingma} optimizer by setting the learning rate as 0.0001.
To alleviate overfitting, we put a \texttt{dropout} layer \cite{srivastava2014dropout} with a rate of 0.5 in the GGF layer and before the output \texttt{softmax} layer, and set the coefficient of $L$2 regularization over the network weights as 0.01.

We adopt two widely used metrics for evaluation: weighted accuracy (WA), which is the overall classification accuracy and unweighted accuracy (UA), which is the average recall over the emotion categories.
We first compute the metrics for each fold and then present the average accuracy over all the folds.

\subsection{Comparison of Representations} \label{sec:representations-compare}
Using the CNN-LSTM encoders, the last hidden state of the bidirectional LSTM layer can be used to represent a speech utterance as $h_s$ and the corresponding text as $h_t$.
Adopting the bidirectional alignment network, we obtained the attention-aligned representations $a_s$ for speech and $a_t$ for text.
We evaluate their discriminative power in emotion classification using the same classification layer (See Section~\ref{sec:softmax}). The accuracy (WA) comparison among the representations on the five folds is shown in Table~\ref{tab:representations}.
\begin{table}[htb]
\centering
\caption{Comparison across representations in emotion classification accuracy (WA) on the IEMOCAP dataset (5 folds).}
\label{tab:representations}
\resizebox{0.7\linewidth}{!}{%
\begin{tabular}{c|cc|cc}
\hline
\textbf{Fold} & $h_s$ & $h_t$ & $a_s$ & $a_t$ \\ \hline
1 & 0.5860 & 0.6285 & 0.6476 & \textbf{0.6762} \\
2 & 0.6384 & 0.7060 & 0.6728 & \textbf{0.7368} \\
3 & 0.5920 & 0.5860 & 0.6150 & \textbf{0.6420} \\
4 & 0.6936 & 0.6494 & 0.6948 & \textbf{0.7289} \\
5 & 0.6030 & 0.6497 & 0.6561 & \textbf{0.6815} \\ \hline
\textbf{Avg} & 0.6226 & 0.6439 & 0.6573 & \textbf{0.6931} \\ \hline
\end{tabular}%
}
\end{table}

We observe that the attention-aligned representations outperform the last-hidden-state representations significantly on all folds for both speech and text.
Since both $a_s$ and $a_t$ take additional information from the other modality, this may explain why they outperform their counterparts.
Another interesting observation is that $a_t$ outperforms all other representations.
This verifies our hypothesis that the alignment direction is important for the learned representation.
The reason that $a_t$ outperforms $a_s$ may be attributed to the different lengths of speech frames and text sequence for the same utterance, where the number of frames can go up to 750 frames while the text sequence may consist of around 20 words.
Since the speech sequence is much longer, the attention-aligned speech representation $a_s$ is not as effective as its counterpart $a_t$.
A similar observation is also reported in \cite{neumann2017attentive}, where the attention mechanism brings slight improvements for speech emotion recognition on the improvised subset of IEMOCAP.

\subsection{Comparison of Information Fusion Methods}
Given different representations obtained from speech and text, the next interesting question to ask is: ``\textit{How to make use of all the representations to achieve better performance?}"
Various information fusion methods can be adopted, such as simple concatenation, tensor fusion network \cite{zadeh2017tensor} (TFL), gated multimodal units \cite{arevalo2017gated} (GMU) and the proposed group gated fusion (GGF) layer.
Performance comparisons among these fusion methods on the IEMOCAP dataset are shown in Table~\ref{tab:fusion}.

On average, GGF obtains the best accuracy and outperforms all other methods on the folds of 2, 4 and 5.
An interesting observation is that both Concat-1 and Concat-2 are strong baselines, although outperformed by TFL and GMU with ($a_s, a_t$) as the input, and further outperformed by GGF given all the four representations ($a_s$, $a_t$, $h_s$, $h_t$).
Since the IEMOCAP dataset is relatively small, the simple \textit{concatenation} method may even have enough capacity to learn the underlying patterns.
GGF computes a \texttt{sigmoid} weight over the non-linear transformation of the representations within each group, which may be the key to its superior performance to all other methods.
\begin{table}[htb]
\centering
\caption{Comparison across fusion methods in emotion classification accuracy (WA) on the IEMOCAP dataset (5 folds).}
\label{tab:fusion}
\resizebox{0.9\linewidth}{!}{%
\begin{tabular}{c|ccc|cc}
\hline
\multirow{2}{*}{\textbf{Fold}} & \multicolumn{3}{c|}{\textbf{Input ($a_s$, $a_t$)}} & \multicolumn{2}{c}{\textbf{Input ($a_s$, $a_t$, $h_s$, $h_t$)}} \\ \cline{2-6}
 & \textbf{Concat-1} & \textbf{TFL} & \textbf{GMU} & \textbf{Concat-2} & \textbf{GGF} \\ \hline
1 & 0.6805 & 0.6900 & 0.6985 & \textbf{0.6996} & 0.6911 \\
2 & 0.7417 & 0.7405 & 0.7552 & 0.7540 & \textbf{0.7737} \\
3 & 0.6600 & 0.6720 & \textbf{0.6890} & 0.6720 & 0.6770 \\
4 & 0.7440 & 0.7440 & 0.7591 & 0.7478 & \textbf{0.7654} \\
5 & 0.6868 & 0.7038 & 0.6975 & 0.7017 & \textbf{0.7123} \\ \hline
\textbf{Avg} & 0.7026 & 0.7101 & 0.7199 & 0.7150 & \textbf{0.7239} \\ \hline
\end{tabular}%
}
\end{table}

\subsection{Comparison with Existing Approaches}

We compare the proposed GBAN model with the existing published approaches in Table~\ref{tab:cv-results}. They all perform 5-fold cross validation in a speaker-independent manner by using four sessions as training set and the whole remaining session as test set.
All the three speech-only models (i.e., CNN-Att \cite{neumann2017attentive}, LSTM-Att \cite{ramet2018context} and Self-Att \cite{tarantino2019self}) adopt attention mechanism based on CNN, LSTM and self-attention, respectively.
In multimodal emotion recognition, we implement all the models marked with * for fair comparisons using the same experimental settings.

\begin{table}[htb]
\centering
\caption{Comparison based on the IEMOCAP dataset (S:Speech, T: Text). All the experiments use 5-fold cross validation and leave one session out as the test set.} 
\label{tab:cv-results}
\resizebox{0.9\linewidth}{!}{%
\begin{tabular}{c|lcc}
\hline
\textbf{Input} & \multicolumn{1}{c}{\textbf{Model}} & \textbf{WA} & \textbf{UA} \\ \hline
\multirow{3}{*}{\textbf{S}} 
 & CNN-Att (Neumann et al.\cite{neumann2017attentive}) & 0.5610 & - \\
 & LSTM-Att (Ramet et al.\cite{ramet2018context}) & 0.6250 & 0.5960 \\
 & Self-Att (Tarantino et al.\cite{tarantino2019self}) & 0.6810 & 0.6380 \\ \hline
\multirow{5}{*}{\textbf{S+T}} 
& Att-LSTM (Xu et al.\cite{xu2019learning})* & 0.6974 & 0.6401 \\
& BiAtt-Concat* & 0.7026 & 0.6561 \\
& BiAtt-TFL (Zadeh et al.\cite{zadeh2017tensor})* & 0.7101 & 0.6627 \\
& BiAtt-GMU (Arevalo et al.\cite{arevalo2017gated})* & 0.7199 & 0.6584 \\
& \textbf{GBAN (This Paper)} & \textbf{0.7239} & \textbf{0.7008} \\ \hline
\end{tabular}%
}
\end{table}

We observe that the multimodal approaches using both speech and text generally outperform the speech-only approaches.
Note that the \textit{Att-LSTM} model, proposed by Xu et al. \cite{xu2019learning} is outperformed by all other models based on bidirectional aligned representations (BiAtt), since it adopts only a unidirectional alignment between speech and text.
Finally, the proposed GBAN model obtains the best performance in terms of both WA and UA on the IEMOCAP dataset.

\subsection{Weights Analysis}
\begin{figure}[htb]
\centering
\begin{tikzpicture}[thick,scale=0.44] 
\begin{axis}[
width=1.6\linewidth,
ybar,
enlargelimits=0.2,
legend style={ font=\large,
legend columns=-1, column sep=0.2cm},
ylabel={Average Weight},
ylabel style={font=\large},
xtick=data,
xticklabels={$a_s$,$a_t$,$h_s$,$h_t$},
xticklabel style={
  yshift=0pt, 
  font=\Large,
},
extra x ticks={1.5,3.5},
extra x tick labels={Group-1, Group-2},
extra x tick style={
  ticklabel style={yshift=-7pt},
  tickwidth=10,
  font=\Large
}
]
\addplot[c1,fill,error bars/.cd,y dir=both,y explicit,]
coordinates{
   (1,0.4781)
   (2,0.5219)
   (3,0.3383)
   (4,0.6617)
};
\addplot[c2,fill,error bars/.cd,y dir=both,y explicit,]
coordinates {
  (1,0.4768)
  (2,0.5232)
  (3,0.2784)
  (4,0.7216)
};
\addplot[c3,fill,error bars/.cd,y dir=both,y explicit,]
coordinates{
   (1,0.4699)
   (2,0.5301)
   (3,0.2890)
   (4,0.7110)
};
\addplot[c4,fill,error bars/.cd,y dir=both,y explicit,]
coordinates {
  (1,0.4844)
  (2,0.5156)
  (3,0.3646)
  (4,0.6354)
};
\addplot[c6,fill,error bars/.cd,y dir=both,y explicit,]
coordinates {
  (1,0.4779)
  (2,0.5221)
  (3,0.3117)
  (4,0.6883)
};
\legend{Fold-1,Fold-2,Fold-3,Fold-4,Fold-5}
\end{axis}
\end{tikzpicture}
\vspace{-0.5em}
\caption{Average weight of each representation in emotion classification on the 5-fold experiments of the GBAN model.}
\label{fig:weights}
\end{figure}
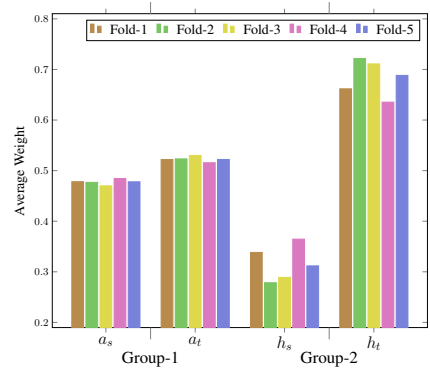

We further investigate the behavior of the proposed GBAN model by analyzing the weights of $z_p$ and $z_q$ in the GGF layer, which determine the contribution of each representation (i.e., $a_s$, $a_t$, $h_s$ and $h_t$) in emotion classification.
Figure~\ref{fig:weights} shows the average weight of each representation for emotion classification on the 5-fold experiments of the GBAN model.
We observe that both $a_t$ and $h_t$ have consistently higher weights within their respective groups (i.e., Group-1 and Group-2) on all the five folds, indicating that the text modality contribute more in emotion classification.
This observation is in line with Section~\ref{sec:representations-compare}, which shows that both $h_t$ and $a_t$ are more discriminative in emotion classification.
Besides, this analysis also illustrates the interpretable benefit of the proposed GGF layer, which allows to present the contribution of a particular modality in multimodal emotion recognition.
\vspace{-0.9em}

\vspace{-1em}
\section{Conclusion}
This paper presents a model named Gated Bidirectional Alignment Network (GBAN) for multimodal emotion recognition, which consists of a novel attention-based bidirectional alignment network to exploit the alignment information between speech and text explicitly, and a new group gated fusion layer to learn the contribution of each representation from both modalities automatically.
We show empirically that the bidirectional alignment network leads to more discriminative representations for emotion classification, and the group gated fusion layer fuses multiple representations effectively in an interpretable manner. GBAN outperforms existing state-of-the-art approaches in emotion classification on the IEMOCAP dataset.

\bibliographystyle{IEEEtran}
\bibliography{references}

\end{document}